\documentclass[sigconf,nonacm]{acmart}
\usepackage{microtype}
\usepackage{graphicx}
\usepackage{subcaption}
\usepackage{booktabs}
\usepackage{svg}
\usepackage{amsmath,mathtools}
\usepackage{amsthm}
\usepackage{algorithm}
\usepackage{algpseudocode}
\usepackage{subfiles}
\usepackage[capitalize,noabbrev]{cleveref}
\usepackage{hyperref}
\usepackage[textsize=tiny]{todonotes}

\title{Context-Aware Initialization for Reducing Generative Path Length in Diffusion Language Models}

\author{Tongyuan Miao}
\affiliation{%
  \institution{University of Michigan}
  \city{Ann Arbor}
  \state{MI}
  \country{USA}
}
\email{tymiao@umich.edu}

\author{Gary Huang}
\affiliation{%
  \institution{University of Michigan}
  \city{Ann Arbor}
  \state{MI}
  \country{USA}
}
\email{ioi@umich.edu}

\author{Kai Jun Han}
\affiliation{%
  \institution{University of Michigan}
  \city{Ann Arbor}
  \state{MI}
  \country{USA}
}
\email{kaijun@umich.edu}

\author{Annie Jiang}
\affiliation{%
  \institution{University of Michigan}
  \city{Ann Arbor}
  \state{MI}
  \country{USA}
}
\email{annij@umich.edu}


\theoremstyle{plain}

\theoremstyle{definition}

\theoremstyle{remark}

\begin{document}

\begin{abstract}
\section{Abstract}
\label{sec:abstract}
Diffusion Large Language Models (DLLMs) enable fully parallel token decoding but often remain impractical at inference time due to the many denoising iterations required to refine an information-free, fully masked initialization into coherent text. Most existing acceleration methods focus on traversing this generative trajectory more efficiently via improved solvers or sampling strategies. We advance a complementary perspective: \emph{shorten the trajectory itself} by starting closer to the target distribution through \emph{context-aware initialization}.

We propose a training-free interface that injects prompt-conditioned priors from a lightweight auxiliary model into the diffusion initialization, and instantiate it with two mechanisms: discrete token injection and representation-level embedding interpolation. Because injected priors can be imperfect and unmask-only decoding can over-commit early, we also introduce a simple confidence-based remasking mechanism as a form of prior skepticism. Preliminary evidence on GSM8K suggests that context-aware initialization can substantially reduce denoising iterations (about 35\% fewer function evaluations in our setting), while also exposing a key open challenge: naive warm-starting can degrade final accuracy relative to strong diffusion baselines. We use these findings to motivate a research agenda around calibration, revision mechanisms, and representation alignment for reliable warm-started diffusion decoding.

\end{abstract}

\maketitle

\section{Introduction}

Consider a simple question-answering prompt:

\begin{verbatim}
Prompt: "2 + 2 = ?"
\end{verbatim}

An autoregressive language model produces the answer ``4'' in a single forward pass. A Diffusion Language Model (DLLM), despite having access to the same context, must instead begin from a fully masked sequence and iteratively refine each token through dozens or even hundreds of denoising steps before ``4'' emerges. This inefficiency persists even when the output is highly predictable: basic arithmetic, common facts, or formulaic responses all demand the same costly iterative refinement.

\begin{figure}[t]
  \centering
  \includegraphics[width=\linewidth]{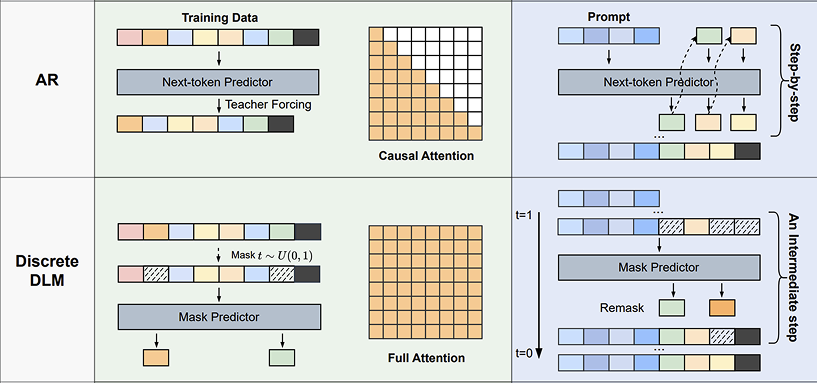}
  \caption{Autoregressive models generate tokens sequentially conditioned on prior outputs, while diffusion language models decode all positions in parallel via iterative denoising. Adapted from \cite{li2025surveydiffusionlanguagemodels}.}
  \label{fig:arvsdiffusion}
\end{figure}

We argue that a primary cause of this inefficiency is \emph{initialization mismatch}. DLLMs begin generation from a state that is extremely far from the target distribution, regardless of how informative the input prompt may be. When a user asks ``The capital of France is\ldots'', the model effectively ``knows'' the answer should be \emph{Paris}, yet the diffusion process forces it to start from a completely masked sequence as if no contextual signal were available. Tokens with high contextual certainty are treated identically to highly uncertain ones.

This mismatch fundamentally limits DLLM efficiency. While diffusion models offer fully parallel token decoding and avoid the sequential bottleneck of autoregressive generation, their practical throughput often falls short of autoregressive systems enhanced with KV-caching. Existing acceleration methods focus primarily on traversing the generative trajectory more efficiently, for example through improved numerical solvers or block-wise unmasking strategies. We propose a complementary perspective: rather than accelerating movement along the diffusion trajectory, we seek to \emph{shorten the trajectory itself} by starting closer to the destination.

\paragraph{Why now.}
Recent progress in diffusion-based language modeling has significantly improved per-step efficiency, but inference latency remains dominated by the number of required denoising iterations. At the same time, modern inference pipelines increasingly rely on auxiliary models—such as autoregressive LLMs, retrievers, or tool-augmented systems—that already produce strong task-conditioned hypotheses. These signals are typically discarded before diffusion begins. This convergence suggests that initialization, rather than solver efficiency alone, represents a natural next bottleneck in diffusion language model inference.

In this paper, we advance the vision of \emph{context-aware initialization} for diffusion language models. Instead of beginning from an information-free masked sequence, we propose conditioning the initial diffusion state on context-dependent signals derived from upstream language models. By injecting semantic structure before denoising begins, the diffusion process can start substantially closer to the final output, reducing the number of required function evaluations while preserving the model’s ability to refine and correct its predictions.

Importantly, our goal is not to present a finalized acceleration technique, but to identify initialization as a critical and underexplored dimension of the DLLM design space. Through a set of training-free warm-start mechanisms and preliminary empirical evidence, we expose both the promise of this approach and the failure modes that must be addressed to make it reliable in practice.

\paragraph{Contributions.}
This paper makes the following contributions:
\begin{itemize}
    \item \textbf{Problem framing.} We identify initialization mismatch as a first-order bottleneck in diffusion language model inference and argue that trajectory length itself is a controllable variable.
    
    \item \textbf{Design space for context-aware initialization.} We introduce a family of training-free warm-start strategies that inject auxiliary model priors into the diffusion initialization, including token-level and embedding-level approaches.
    
    \item \textbf{Correction mechanisms.} We propose remasking as a general principle for revising unreliable warm-start signals during diffusion, enabling the model to reconsider low-confidence initializations.
    
    \item \textbf{Preliminary evidence and insights.} Initial experiments suggest that warm-starting can substantially reduce denoising iterations, while also revealing key challenges related to prior miscalibration and representation mismatch.
\end{itemize}

Together, these contributions motivate a research agenda centered on reliable, context-aware initialization as a path toward making fully parallel diffusion language models practical at scale.

\begin{figure}[t]
  \centering
  \includegraphics[width=\linewidth]{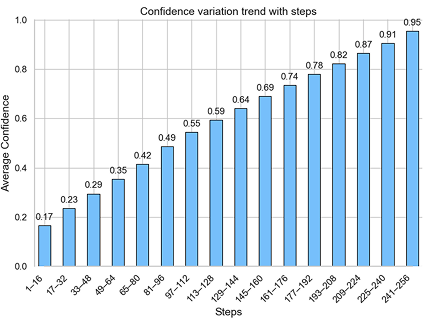}
  \caption{Model confidence increases with denoising steps, suggesting that earlier access to reliable structure could reduce the required refinement budget. Adapted from \cite{nie2025largelanguagediffusionmodels}.}
  \label{fig:confidenceincreasing}
\end{figure}

\section{System Architecture}
\label{sec:arch}

\paragraph{Setting and notation.}
Let the input prompt be a token sequence $p=(p_1,\ldots,p_m)$ and let the desired output length be $n$.
We generate an output sequence $x_0=(x_{0,1},\ldots,x_{0,n})\in\mathcal{V}^n$ over vocabulary $\mathcal{V}$ using a \emph{masked} diffusion language model (DLLM).
The diffusion state at reverse step $t$ is a partially specified sequence
\[
x_t \in (\mathcal{V}\cup\{\texttt{[MASK]}\})^n,
\]
where $\texttt{[MASK]}$ denotes the special mask token and indices $i\in\{1,\ldots,n\}$ refer to output positions.

\paragraph{Baseline initialization.}
In masked diffusion, inference begins from an information-free terminal state in which every position is masked:
\begin{equation}
x_T \;=\; (\texttt{[MASK]},\ldots,\texttt{[MASK]}) \in (\mathcal{V}\cup\{\texttt{[MASK]}\})^n.
\label{eq:all_mask_init}
\end{equation}
Because $x_T$ contains no lexical content, early reverse steps must discover coarse structure from scratch, which can increase the number of denoising iterations required to reach a high-quality $x_0$.

\paragraph{Goal and high-level approach.}
We seek to reduce inference cost by \emph{warm-starting} the reverse process: we replace Eq.~\eqref{eq:all_mask_init} with a prompt-conditioned initialization that injects semantic structure \emph{without retraining} the DLLM.
Concretely, we compute a prompt-conditioned warm proposal using a smaller auxiliary language model, then inject this signal into the DLLM initialization either as discrete token IDs (token-level prior) or as continuous embeddings (representation-level prior).

\paragraph{Design requirements and scope.}
We focus on a training-free interface that (i) modifies only inference-time initialization, (ii) preserves compatibility with existing Fast-DLLM decoding (e.g., confidence-threshold unmasking), and (iii) retains the ability to revise unreliable priors during generation. We do not claim that the specific instantiations below are optimal; rather, they serve as concrete points in a broader design space for context-aware initialization.

\paragraph{Warm proposal generator.}
We use an auxiliary autoregressive model $f_\phi$ to produce prompt-conditioned token proposals
\begin{equation}
\hat{x} = f_\phi(p),
\label{eq:aux_model}
\end{equation}
where $\hat{x}=(\hat{x}_1,\ldots,\hat{x}_n)$ and $\hat{x}_i\in\mathcal{V}$.

\paragraph{Diffusion-embedding warm lookup.}
Let $\mathrm{Emb}_\theta(\cdot)$ denote the DLLM token embedding layer. Given warm token IDs $\hat{x}$, define
\begin{equation}
\hat{e}_i = \mathrm{Emb}_\theta(\hat{x}_i), \qquad i=1,\ldots,n,
\label{eq:warm_embed_lookup}
\end{equation}
and collect them into $\hat{E}=(\hat{e}_1,\ldots,\hat{e}_n)\in\mathbb{R}^{n\times d}$.

\paragraph{Warm initialization operator.}
We define an operator $\mathcal{W}$ that combines the baseline all-mask state and warm proposals to produce a warm-started terminal condition:
\begin{equation}
\tilde{x}_T \;\;\text{or}\;\; \tilde{E}_T \;=\; \mathcal{W}(x_T,\hat{x},\hat{E}).
\label{eq:warm_operator}
\end{equation}
We instantiate $\mathcal{W}$ in two ways:
(i) direct replacement of a subset of masked positions with warm token IDs (Sec.~\ref{sec:method1}), and
(ii) keeping $x_T$ fully masked while warm-starting the \emph{input representations} via embedding interpolation (Sec.~\ref{sec:method2}). In the latter case, $\tilde{E}_T$ is fed as the initial embedding sequence to the DLLM in place of $\mathrm{Emb}_\theta(x_T)$.

\paragraph{Diffusion backbone (unchanged).}
Let $p_\theta(x_{t-1}\mid x_t,p)$ denote the DLLM reverse transition distribution for $t=T,\ldots,1$.
Our approach does not modify $p_\theta$ or its parameters; all changes occur at inference time through $(\tilde{x}_T \text{ or } \tilde{E}_T)$ and (optionally) mechanisms that reduce or revoke prior influence as denoising progresses.

\section{Methods}
\label{sec:methods}

\subsection{Background: confidence-threshold unmasking}
\label{sec:unmask_background}

We adopt the confidence-aware decoding strategy from Fast-dLLM \cite{wu2025fastdllmtrainingfreeaccelerationdiffusion}.
Starting from a partially masked sequence, decoding proceeds in iterations. At each iteration, the diffusion model produces per-position logits for masked positions; we convert logits to probabilities and compute a confidence score (e.g., max softmax probability). All positions whose confidence exceeds a threshold $\tau$ are unmasked in parallel; if none exceed $\tau$, we unmask the single most confident position to guarantee progress. As in Fast-dLLM, the baseline is \emph{unmask-only} for model-decoded tokens: once a token is revealed, it is kept fixed in later iterations.

\subsection{Warm-start design space: injecting priors at initialization}
\label{sec:warmstart_methods}

This section specifies two instantiations of the warm initialization operator $\mathcal{W}$ (Sec.~\ref{sec:arch}). Both are training-free and modify only inference-time initialization, but they inject auxiliary information at different interfaces: the discrete token state vs.\ the continuous representation state.

\subsubsection{Token-level priors: Method 1 (Token injection)}
\label{sec:method1}

\begin{figure}[t]
  \centering
  \includegraphics[width=\linewidth]{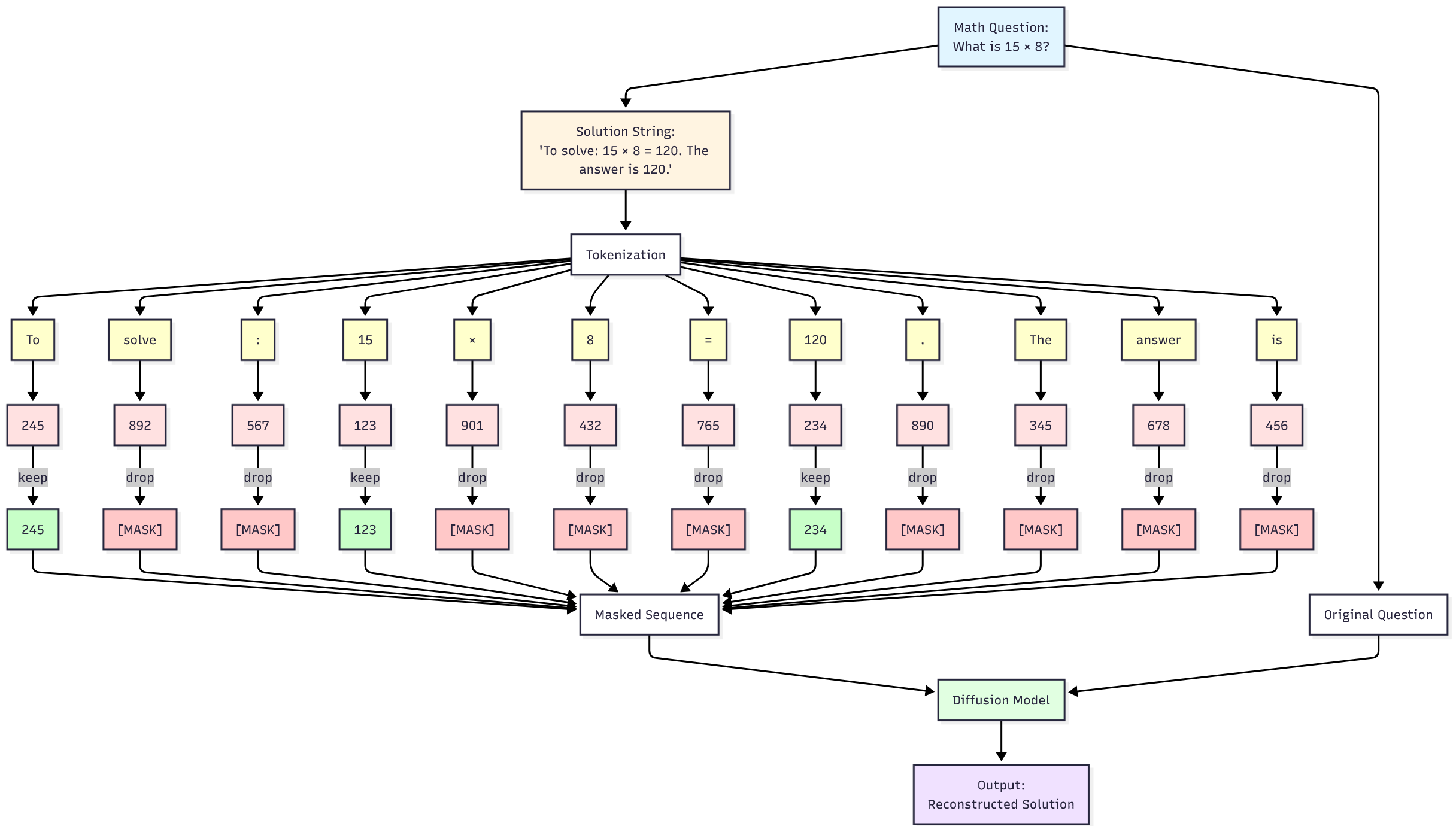}
  \caption{Token injection: initialize a subset of positions with auxiliary warm tokens (injection dropout), leaving the rest masked. Injected tokens act as a noisy prior and can be revoked by remasking.}
  \label{fig:method1}
\end{figure}

We initialize the discrete state by injecting auxiliary warm tokens into a subset of positions and leaving the rest masked.
Let $g_i\sim\mathrm{Bernoulli}(\rho)$ be an injection gate with rate $\rho\in[0,1]$. The initialized token at position $i$ is
\begin{equation}
x^{(0)}_i =
\begin{cases}
\hat{x}_i & \text{if } g_i=1,\\
\texttt{[MASK]} & \text{if } g_i=0,
\end{cases}
\qquad i\in[n].
\label{eq:method1_init}
\end{equation}
Let $I \triangleq \{i\in[n] : g_i=1\}$ denote injected positions.

\paragraph{Interpretation and failure mode.}
Token injection provides a strong, discrete prior that can sharply reduce the search space early in denoising. However, when injected tokens are incorrect, unmask-only decoding can over-commit and propagate early mistakes. This motivates explicit revision mechanisms (Sec.~\ref{sec:remasking}) and suggests that \emph{calibration of prior trust} is a central design question.

\subsubsection{Representation-level priors: Method 2 (Embedding interpolation)}
\label{sec:method2}

\begin{figure}[t]
  \centering
  \includegraphics[width=\linewidth]{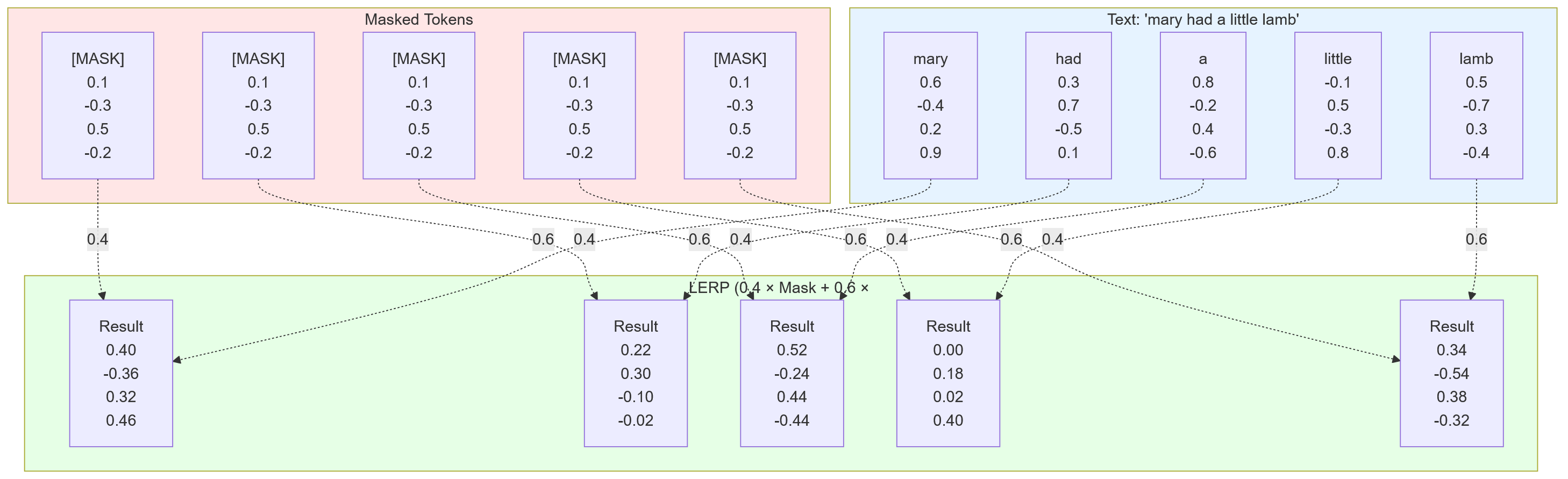}
  \caption{Embedding interpolation: keep the discrete initialization fully masked, but warm-start the DLLM by interpolating the mask embedding with the warm-token embedding, with injection dropout applied in embedding space.}
  \label{fig:method2}
\end{figure}

Method~2 keeps the discrete initialization fully masked, but biases the initial \emph{representations} toward the auxiliary warm proposal.
Let $e_{\text{mask}}=\mathrm{Emb}_\theta(\texttt{[MASK]})$ and let $\hat{e}_i=\mathrm{Emb}_\theta(\hat{x}_i)$ be defined as in Eq.~\eqref{eq:warm_embed_lookup}.
We form the interpolated embedding
\begin{equation}
\tilde{e}^{(0)}_i \;=\; (1-\alpha)\,e_{\text{mask}} + \alpha\,\hat{e}_i,
\qquad \alpha\in[0,1],
\label{eq:method2_interp}
\end{equation}
where $\alpha$ controls the strength of the representation-level prior.
To prevent over-reliance on the auxiliary signal, we apply injection dropout in embedding space:
\begin{equation}
e^{(0)}_i \;=\;
\begin{cases}
\tilde{e}^{(0)}_i & \text{with probability } \rho,\\
e_{\text{mask}} & \text{with probability } 1-\rho.
\end{cases}
\label{eq:method2_dropout}
\end{equation}

\paragraph{Interpretation and open question.}
Embedding interpolation provides a \emph{soft} prior while leaving the discrete sequence fully revisable. A key open question is whether interpolated embeddings remain on-manifold for the DLLM’s learned representation space; if not, the model may receive a misleading signal even when $\hat{x}$ is partially correct. This motivates future work on representation alignment (e.g., learned projections, normalization, or adaptive $\alpha$ schedules) that preserve revisability without introducing off-manifold artifacts.

\subsection{Prior skepticism via remasking}
\label{sec:remasking}

Fast-dLLM decoding is unmask-only: once a token is revealed, it is not revised \cite{wu2025fastdllmtrainingfreeaccelerationdiffusion}. Under context-aware initialization, this creates a tension: strong priors can shorten trajectories, but incorrect priors can cause early error lock-in. We therefore treat \emph{remasking} as a simple form of \emph{prior skepticism}: the model may revoke low-confidence injected tokens and reconsider them later in denoising.

\paragraph{Current-iteration token confidence.}
At decoding iteration $k$, the diffusion model produces a logit vector for each position $i$:
\begin{equation}
\ell_i^{(k)} \in \mathbb{R}^{|\mathcal{V}|}.
\end{equation}
We convert logits to probabilities with
\begin{equation}
\pi_i^{(k)}(v) \triangleq \mathrm{softmax}(\ell_i^{(k)})_v, \qquad v\in\mathcal{V},
\label{eq:softmax_probs}
\end{equation}
and define confidence in the \emph{currently fixed} token $x_i^{(k)}$ as
\begin{equation}
\bar{c}_i^{(k)} \triangleq \pi_i^{(k)}\!\left(x_i^{(k)}\right).
\label{eq:current_token_conf}
\end{equation}

\paragraph{A simple decaying remask rule (baseline instantiation).}
We implement one straightforward instantiation where remasking probability depends on current uncertainty and a decaying bias that discourages remasking late in decoding. For an injected position $i\in I$, define a linearly decaying bias
\begin{equation}
b^{(k)} \;=\; b_0 \;-\; \lambda k,
\qquad b_0>0,\; \lambda>0,
\label{eq:linear_bias}
\end{equation}
and a per-token remasking rate
\begin{equation}
r_i^{(k)} \;\triangleq\; \mathrm{clip}_{[0,1]}\!\left(\bigl(1-\bar{c}_i^{(k)}\bigr) + b^{(k)}\right).
\label{eq:remask_rate}
\end{equation}
We then sample a remask decision independently:
\begin{equation}
\delta_i^{(k)} \sim \mathrm{Bernoulli}\!\left(r_i^{(k)}\right), \qquad
x_i^{(k)} \leftarrow \texttt{[MASK]} \ \ \text{if}\ \ \delta_i^{(k)}=1.
\label{eq:remask_sampling}
\end{equation}

\paragraph{Design levers.}
We emphasize that remasking schedules (linear vs.\ nonlinear decay), per-token gating, and confidence calibration are open design dimensions. More generally, ``when to trust the prior'' vs.\ ``when to revoke it'' is a central question for context-aware initialization and a natural target for future work.

\paragraph{Progress guarantee.}
We preserve Fast-dLLM's progress guarantee by ensuring that at least one masked token is unmasked each iteration (even if no position exceeds $\tau$), preventing infinite loops \cite{wu2025fastdllmtrainingfreeaccelerationdiffusion}.

\subsection{Unified inference}
\label{sec:unified}

Given a prompt $p$ and target length $n$, inference proceeds in three stages.

\textbf{(1) Warm-start construction:} we first generate a prompt conditioned proposal $\hat{x}=f_\phi(p)$ (Eq.~\eqref{eq:aux_model}) and then form the initial condition via $\mathcal{W}$ (Eq.~\eqref{eq:warm_operator}). Method~1 produces a partially specified discrete state $\tilde{x}_T$ by token injection (Eq.~\eqref{eq:method1_init}); Method~2 keeps the discrete sequence fully masked but replaces the initial embedding sequence with $\tilde{E}_T$ via interpolation and dropout (Eqs.~\eqref{eq:method2_interp}--\eqref{eq:method2_dropout}).

\textbf{(2) Parallel decoding:} starting from $(\tilde{x}_T \text{ or } \tilde{E}_T)$, we run confidence-threshold unmasking as in Fast-dLLM \cite{wu2025fastdllmtrainingfreeaccelerationdiffusion}. At each iteration $k$, the DLLM produces logits $\ell_i^{(k)}$ for masked positions, which are converted to probabilities $\pi_i^{(k)}$ (Eq.~\eqref{eq:softmax_probs}). Positions whose confidence exceeds $\tau$ are unmasked in parallel; if none exceed $\tau$, we unmask the single most confident position to guarantee progress.

\textbf{(3) Optional correction:} for injected positions $i\in I$, we compute current-token confidence $\bar{c}_i^{(k)}$ (Eq.~\eqref{eq:current_token_conf}) and stochastically remask according to Eqs.~\eqref{eq:linear_bias}--\eqref{eq:remask_sampling}. Decoding terminates when no masked tokens remain, yielding the final sequence $x_0$.

\section{Preliminary Evidence}
\label{sec:experiments}

This section provides \emph{preliminary evidence} that context-aware initialization can shorten diffusion trajectories by reducing the number of denoising iterations required for convergence. Our intent is not to claim finalized end-to-end acceleration, but to (i) demonstrate that initialization is a controllable axis of DLLM efficiency, and (ii) surface the key failure modes that must be addressed to make warm-starting reliable.

\paragraph{NFE vs. end-to-end latency.}
Throughout, we report the Number of Function Evaluations (NFE) as an algorithmic proxy for diffusion compute. However, NFE does not capture total wall-clock latency, which must include the overhead of generating warm-start signals from an auxiliary model. Measuring end-to-end throughput and exploring amortization strategies (e.g., lightweight proposal models, caching, or reuse across generations) is an important next step and is deferred here.

\subsection{Experimental Setup}

\textbf{Models.} We use the \textbf{Fast-dLLM} framework~\cite{wu2025fastdllmtrainingfreeaccelerationdiffusion} as our diffusion backbone, specifically the \texttt{GSAI-ML/LLaDA-8B-Instruct} checkpoint. Warm-start signals are generated using two auxiliary autoregressive models: \texttt{meta-llama/Llama-2-7b-hf} and \\ \texttt{meta-llama/Llama-3.1-8B-Instruct}. We compare against: (i) autoregressive LLaMA-2 and LLaMA-3 generation, and (ii) Fast-dLLM with standard mask initialization.

\textbf{Dataset and metrics.} We evaluate on \textbf{GSM8K}~\cite{cobbe2021training} using a randomly sampled subset of 128 test examples due to computational constraints. We report:
\begin{itemize}
    \item \textbf{Average NFE:} average denoising iterations to convergence (lower is shorter trajectory).
    \item \textbf{Strict-Match Accuracy:} exact match to the ground-truth answer.
    \item \textbf{Flexible-Extract Accuracy:} extracted numeric answer correctness (format-tolerant).
\end{itemize}

\textbf{Implementation details.} Experiments were conducted on NVIDIA RTX 4090 and A40 GPUs, with results primarily derived from RTX 4090 runs. We integrate initialization strategies into \\ \texttt{lm-evaluation-harness}. For Method 1 (Token Injection) we use injection rate $\rho = 0.25$. For Method 2 (Embedding Interpolation) we use interpolation weight $\alpha = 0.6$ and $\rho = 0.25$.

\subsection{Main Observations}

Table~\ref{tab:main_results} summarizes results. Overall, warm-starting substantially reduces NFE, but introduces a pronounced quality gap relative to the strong Fast-dLLM baseline. We treat this gap not as a negative conclusion, but as evidence that \emph{prior reliability and representation alignment} are central challenges for context-aware initialization.

\begin{table*}[t]
\centering
\caption{Preliminary results on GSM8K (128 samples). Context-aware initialization reduces average denoising iterations (NFE) relative to Fast-dLLM, while highlighting a quality gap that motivates correction and calibration mechanisms.}
\label{tab:main_results}
\renewcommand{\arraystretch}{1.3} 
\setlength{\tabcolsep}{12pt} 
\begin{tabular}{lccc}
\toprule
\textbf{Model / Method} & \textbf{Avg NFE} $\downarrow$ & \textbf{Strict-Match} $\uparrow$ & \textbf{Flex-Extract} $\uparrow$ \\
\midrule
\textit{Autoregressive Baselines} & & & \\
LLaMA-2 (7B) & N/A & 0.1069 & 0.1160 \\
LLaMA-3 (8B Instruct) & N/A & \textbf{0.6111} & 0.6899 \\
\midrule
\textit{Diffusion Baselines} & & & \\
Original Fast-dLLM & 79.12 & 0.3594 & \textbf{0.7812} \\
\midrule
\textit{Ours (Context-Aware)} & & & \\
Method 1 (Token Injection) & \textbf{51.70} & 0.2266 & 0.2656 \\
Method 2 (Embedding Interp.) & 56.40 & 0.2500 & 0.2656 \\
\bottomrule
\end{tabular}
\end{table*}

\textbf{Trajectory shortening (NFE reduction).}
Fast-dLLM requires 79.12 denoising steps on average. With context-aware initialization, Method 1 reduces this to 51.70 steps (approximately 35\% fewer iterations), and Method 2 reduces it to 56.40 steps. These results provide initial evidence that initialization alone can meaningfully shorten the diffusion trajectory, independent of solver changes.

\textbf{Quality degradation as a diagnostic signal.}
Both context-aware methods underperform the Fast-dLLM baseline in accuracy, suggesting that injected priors can ``lock in'' early mistakes or bias the diffusion dynamics toward incorrect hypotheses. At the same time, both methods improve substantially over the weak LLaMA-2 proposal model (e.g., Flex-Extract 0.2656 vs.\ 0.1160), indicating that diffusion refinement can correct some errors from a weak prior. The remaining gap to standard initialization highlights the need for better mechanisms to (i) detect unreliable priors and (ii) preserve the model’s ability to revise early commitments.

\subsection{Analysis: What the Results Suggest}

\textbf{Token vs.\ embedding priors.}
Method 1 (hard token injection) and Method 2 (soft embedding interpolation) achieve similar accuracy, with Method 2 trading slightly higher NFE for slightly higher Strict-Match. This suggests that merely providing a prior signal is insufficient: the \emph{interface} between auxiliary representations and the DLLM’s denoising dynamics likely determines whether warm-starting helps or harms. A key open question is how to inject soft priors while keeping them on-manifold for the DLLM’s learned representation space.

\textbf{Correction mechanisms remain an open lever.}
We implemented confidence-aware remasking intended to revoke low-confidence warm-start tokens, but did not fully tune or isolate its contribution in the current study. In a vision framing, we view remasking schedules and calibration as first-class research directions: how should the model decide when to trust the prior, when to revoke it, and how aggressively to revisit early decisions?

\subsection{Summary and Open Questions}

In summary, these preliminary experiments show that context-aware initialization can substantially reduce diffusion iteration count on GSM8K, while also exposing a central challenge: naive warm-starting can degrade accuracy relative to strong diffusion baselines. This motivates several immediate research questions: (i) how to calibrate confidence under injected priors, (ii) how to design remasking or revision mechanisms that prevent early error lock-in, and (iii) how to align soft priors with the DLLM representation manifold (see, e.g.,~\cite{hersche2025softmaskeddiffusionlanguagemodels}).

\section{Related Works}
\label{sec:related_works}

\subsection{Diffusion Language Models and Efficiency}
Discrete diffusion models for language have recently closed much of the quality gap to autoregressive (AR) approaches. Score Entropy Discrete Diffusion (SEDD) introduces a discrete diffusion objective (\emph{score entropy}) and reports competitiveness with AR language models on standard language modeling benchmarks, with the ability to trade compute for quality and support controllable infilling \cite{lou2024discretediffusionmodelingestimating}. More recently, LLaDA scales masked diffusion to an instruction-tuned setting by training a diffusion language model from scratch and reports competitive in-context learning performance relative to strong AR baselines, with additional qualitative capabilities after supervised fine-tuning \cite{nie2025largelanguagediffusionmodels}. 

Despite these advances, diffusion language models remain burdened by iterative sampling: generation requires many denoising iterations (i.e., many network evaluations) to refine an initially uninformative masked sequence into a coherent output. This iterative cost motivates work on improving inference efficiency without sacrificing output quality. A representative direction is \emph{block diffusion}: BD3-LMs interpolate between autoregressive and diffusion modeling by denoising within blocks, enabling flexible-length generation and improving inference efficiency via KV caching and parallel token sampling \cite{arriola2025blockdiffusioninterpolatingautoregressive}.

\subsection{Optimizing the Generative Trajectory}
A major line of research focuses on accelerating traversal of the diffusion trajectory via improved sampling/decoding strategies. Fast-dLLM is a training-free acceleration framework that enables KV caching and parallel decoding for diffusion LLMs, reducing inference cost without modifying model weights \cite{wu2025fastdllmtrainingfreeaccelerationdiffusion}. More recent work adapts decoding to task structure and error dynamics. Saber proposes a training-free sampling algorithm for code generation motivated by two observations: diffusion decoding can be adaptively accelerated as more code context is established, and effective sampling benefits from backtracking mechanisms to reverse generated tokens when needed \cite{dong2025saberefficientsamplingadaptive}. ReMDM introduces a remasking diffusion model sampler that enables iterative refinement by allowing previously generated tokens to be updated again, and emphasizes inference-time scaling as additional compute is allocated \cite{wang2025remaskingdiscretediffusionmodels}. 

Notably, these sampling-focused approaches generally operate within the standard masked-diffusion paradigm in which generation begins from a fully masked (uninformative) initial state, and the main optimization target is how efficiently the model traverses the resulting denoising path.

\subsection{Warm-Starting and Hybrid Generation}
A complementary perspective is to shorten the generative path by altering the initialization state. In the image domain, MoDM is a caching-based serving system for diffusion models that reuses cached final images to reduce serving time while preserving image quality, dynamically balancing latency and quality through a mixture of diffusion models \cite{xia2025modmefficientservingimage}. While MoDM is a deployment-oriented system rather than a language-generation method, it highlights the general advantage of starting the reverse process from a semantically meaningful anchor rather than from an information-free initialization.

Our work translates the warm-start intuition to diffusion language models using a generative (rather than retrieval-first) mechanism: a lightweight auxiliary autoregressive model produces a prompt-conditioned proposal, which we inject as a noisy prior into the diffusion initialization. This is conceptually related to hybrid AR+non-AR decoding ideas, but here the proposal is used to shape the diffusion starting point and is refined (and potentially overwritten) through diffusion updates. Because injected priors can be imperfect, we discuss correction mechanisms inspired by the broader remasking/backtracking literature (ReMDM and Saber) while focusing on the distinct setting where the initial tokens originate from an external auxiliary model \cite{wang2025remaskingdiscretediffusionmodels, dong2025saberefficientsamplingadaptive}.

\section{Conclusion}
\label{sec:conclusion}

This paper argues that \emph{initialization is a major bottleneck} in masked diffusion decoding. While prior work largely optimizes how efficiently diffusion traverses the denoising path, we propose shortening the path itself by starting from prompt-conditioned priors. We explored two training-free warm-start interfaces---token injection and embedding-level interpolation---and a simple remasking mechanism that enables revision of unreliable priors during decoding.

Our preliminary evidence suggests that warm-starting can substantially reduce denoising iterations, but also exposes a key failure mode: externally injected priors can lead to early commitment under unmask-only decoding, producing a quality gap relative to strong diffusion baselines. At the same time, the diffusion model can refine weak priors into improved outputs, indicating that warm-starting is not merely copying but a potentially powerful form of hypothesis refinement. We view these results as a diagnostic that motivates a broader research agenda around calibration, revision, and representation alignment for reliable context-aware initialization.
\section{Future Work}
\label{sec:future_work}

Our results suggest several concrete directions for making context-aware initialization reliable:

\paragraph{Calibration and trust in injected priors.}
Warm-starting changes the distribution of early decoding states. A central question is how to calibrate confidence thresholds and token reliabilities under injected priors, and how to detect when the proposal is systematically misleading.

\paragraph{Revision policies and remasking schedules.}
We treated remasking as a simple form of prior skepticism, but the space of revision policies is broad. Understanding when to revoke tokens, how aggressively to backtrack, and how revision interacts with confidence-threshold unmasking \cite{wu2025fastdllmtrainingfreeaccelerationdiffusion} are key steps toward closing the accuracy gap without sacrificing iteration reductions.

\paragraph{Representation alignment for soft warm-starts.}
The comparable performance of embedding interpolation and token injection suggests that na\"ive interpolations may be off-manifold for models trained with a single \texttt{[MASK]} embedding. Developing on-manifold soft masking schemes and lightweight adaptation procedures may be necessary to realize the promise of representation-level priors \cite{hersche2025softmaskeddiffusionlanguagemodels}.

\paragraph{End-to-end throughput and broader evaluation.}
Finally, reducing NFE must translate into wall-clock gains after proposal overhead, caching, and batching effects are accounted for. Broader evaluations across tasks and controlled analyses of quality over iterations will help identify where initialization provides the largest practical benefit and where it fails.

\bibliographystyle{ACM-Reference-Format}
\bibliography{references}

\end{document}